%% file: main.tex
\documentclass[10pt,twocolumn,letterpaper]{article}

\usepackage[pagenumbers]{cvpr}   

\input{preamble}

\definecolor{cvprblue}{rgb}{0.21,0.49,0.74}
\usepackage[pagebackref,breaklinks,colorlinks,allcolors=cvprblue]{hyperref}

\def\confName{CVPR}
\def\confYear{2027}

\title{SIR-Agent: Skill-Augmented Image Restoration Agent with Recursive Self-Improvement}

\author{
Yijian Wang$^{1}$ \qquad
Qingsen Yan$^{1,2}$ \qquad
Weiyu Zhou$^{1}$ \qquad
Mingde Zhang$^{1}$ \\
Axi Niu$^{1}$ \qquad
Haokui Zhang$^{1}$ \qquad
Jiantao Zhou$^{3}$ \\
$^{1}$Northwestern Polytechnical University\\
$^{2}$Shenzhen Research Institute of Northwestern Polytechnical University\\
$^{3}$University of Macau\\
{\tt\small wangyijian1017@163.com, \{qingsenyan,nax,hkzhang\}@nwpu.edu.cn}\\
{\tt\small weiyvzhou@gmail.com, 2025302498@mail.nwpu.edu.cn, jtzhou@um.edu.mo}
}

\begin{document}

\maketitle
\input{sec/0_abstract}    
\input{sec/1_intro}
\input{sec/2_related}

\input{sec/3_method}
\input{sec/4_experiment}

\input{sec/5_conclusion}

{
    \small
    \bibliographystyle{ieeenat_fullname}
    \bibliography{main}
}


\end{document}

%% file: preamble.tex
\usepackage{colortbl,multirow}
\usepackage{tabularx}
\definecolor{irTableHeader}{RGB}{255,204,199}

\usepackage[resetlabels]{multibib}
\newcites{supp}{References}

\makeatletter
\AtBeginDocument{%
    \patchcmd{\Hy@backout}{\@auxout}{\@mainaux}{}{\PackageError{supprefs}{Backref patch failed}{}}
    \patchcmd{\Hy@backout}{\@auxout}{\@mainaux}{}{\PackageError{supprefs}{Backref patch failed}{}}
}
\makeatother

\makeatletter
\AtEndPreamble{%
  \@ifpackageloaded{lineno}{%
    \patchcmd{\logtheLineNumber}{\@auxout}{\@mainaux}{}{\PackageError{supprefs}{Line number cache patch failed}{}}
    \patchcmd{\WriteLineNo}{\@auxout}{\@mainaux}{}{\PackageError{supprefs}{Line label cache patch failed}{}}
    \RemoveFromHook{build/column/after}[lineno]
    \AddToHook{build/column/after}[lineno]{%
        \setbox\@outputbox\vbox to\ht\@outputbox{%
            \boxmaxdepth\@maxdepth
            \protected@write\@mainaux{}{%
                \string\def\string\@LN@column{\if@firstcolumn1\else2\fi}%
            }%
            \unvbox\@outputbox
        }%
    }%
  }{}%
}
\makeatother



%% file: sec/0_abstract.tex
\begin{abstract}
Agents for Image Restoration (IR) aim to leverage Multimodal Large Language Models (MLLMs) to autonomously perceive image degradation and invoke IR tools. 
However, existing IR agents typically rely on fine-tuning MLLM parameters to adapt to new IR tasks, leading to catastrophic forgetting of IR knowledge. 
To address this limitation, we propose a skill-augmented IR agent, dubbed SIR-Agent, that maintains IR knowledge in a scalable  skill library and uses Recursive Self-Improvement (RSI) to facilitate efficient skill-library evolution. 
Specifically, we design a Chain Feedback-driven Skill Revision (CFSR) to evolve the skill library. 
The CFSR analyzes restoration chains to identify and revise weak operations in existing skills. 
The compatibility of a revised skill with known tasks decides if it replaces the original skill or is appended independently. 
Furthermore, to advance CFSR’s skill-library evolution, we present an RSI-driven skill meta-learning mechanism, which identifies CFSR’s inefficient operations and autonomously refines its revision strategy. 
Extensive experiments across 14 IR benchmarks, covering 6 single-degradation and 8 mixed-degradation scenarios, demonstrate SIR-Agent's superior IR performance. 
Further evaluation on 7 unseen tasks validates SIR-Agent's generalization to new IR tasks.
\end{abstract}

%% file: sec/1_intro.tex
\section{Introduction}
\label{sec:intro}

Real-world images often contain multiple interacting degradations, making image restoration (IR) a problem of coordinating restoration operations.
Task-specific restoration models~\cite{zamir2021multistage,zamir2022restormer,wang2022uformer,guo2025mambairv2} have improved restoration for individual degradations, while all-in-one models~\cite{potlapalli2023promptir,qin2024restore,wu2026beyond,wu2026gradient} support multiple restoration tasks within a shared network.
For complex inputs, restoration quality also depends on which tools are selected, their execution order, and their configurations, since each operation changes the image presented to subsequent tools~\cite{chen2024restoreagent}.
Multimodal large language model (MLLM)-driven IR agents address these decisions by perceiving degradation and invoking appropriate tools, allowing the restoration process to adapt to the input~\cite{chen2024restoreagent,zhu2025intelligent,zhou2026qagent,wei2026iamagent,kong2026harnessir}.
Their effectiveness therefore depends on restoration strategy knowledge: when an operation is useful, how it interacts with other operations, and how the knowledge can guide future tasks.

\begin{figure}[!t]
  \centering
  \includegraphics[width=\linewidth]{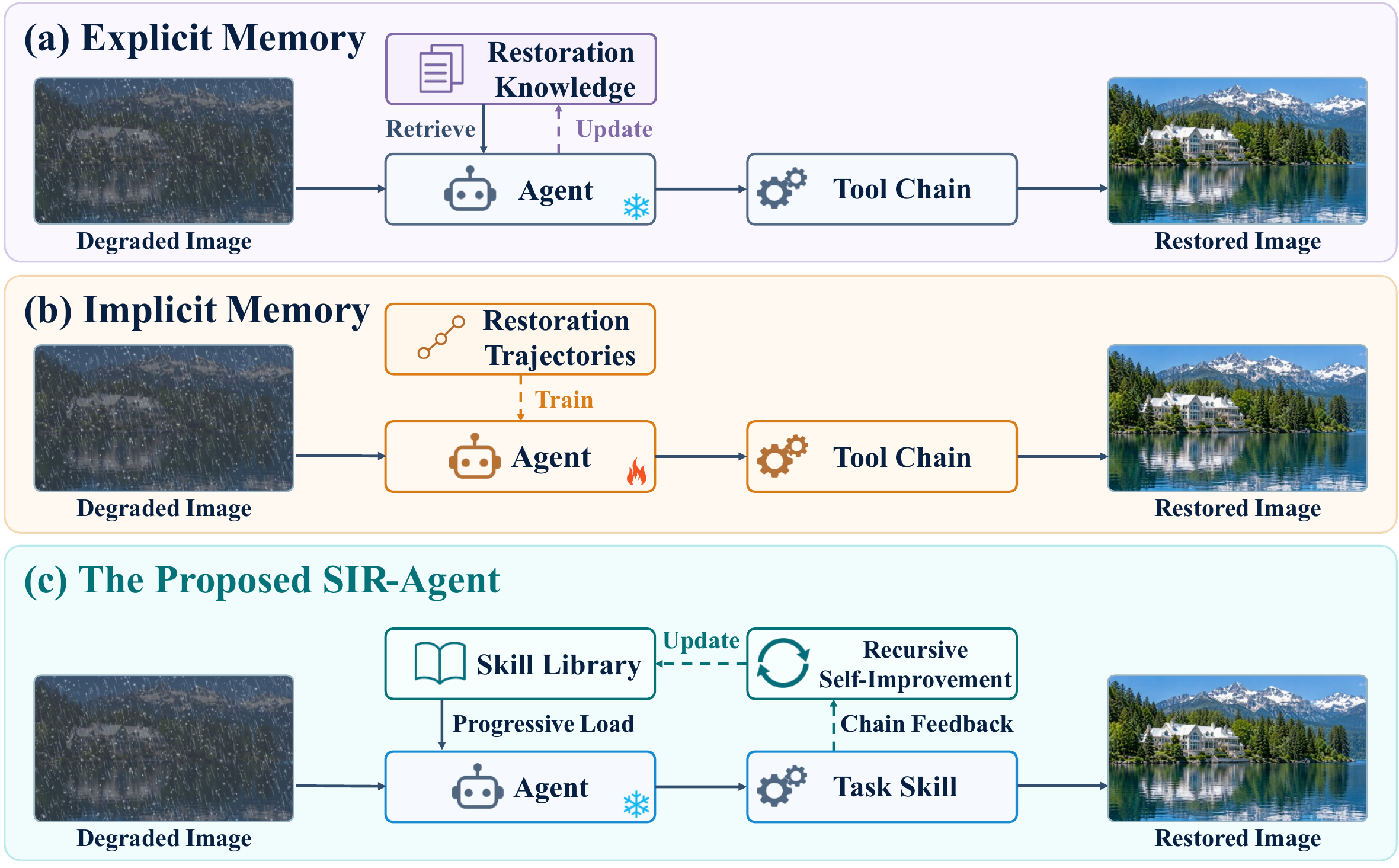}
  \caption{Execution workflows of existing IR agents and our SIR-Agent. 
  (a) Explicit-memory methods retrieve external restoration knowledge for tool-chain generation.
  (b) Implicit-memory methods encode restoration knowledge into model parameters via training.
  (c) The SIR-Agent progressively loads skills with a frozen MLLM. Chain feedback enables skill revision, while RSI optimizes revision rules.}
  \label{fig:motiva}
\end{figure}

Existing IR agents~\cite{chen2024restoreagent,zhu2025intelligent,jiang2026multiagent,li2026hybrid,lu2026restorer1,zuo2025fourkagent,lin2025jarvisir,wang2026paagent,jia2026tiragent,cui2026selfevolving,zhu2026opera,choi2026ditto,zhuang2026evoiragent,gao2026causalagentir,zhang2026look,xiong2026beyond,yu2026derainagent,zhu2026epiagent,tang2026evolutionary,liu2025realworld,zhou2026hdragent} acquire and use restoration strategy knowledge through two complementary approaches, illustrated in Fig.~\ref{fig:motiva}.
Explicit memory-based IR agents~\cite{zhu2025intelligent,jiang2026multiagent,zuo2025fourkagent,wang2026paagent,gao2026causalagentir,zhuang2026evoiragent,cui2026selfevolving,tang2026evolutionary,zhu2026epiagent,zhou2026hdragent} store restoration strategy knowledge outside the MLLM parameters.
As shown in Fig.~\ref{fig:motiva}(a), the agent retrieves external restoration strategy knowledge to generate a tool chain, while restoration feedback supports updates to the external memory.
However, these agents ~\cite{zhu2025intelligent,jiang2026multiagent,zuo2025fourkagent,wang2026paagent,gao2026causalagentir,zhuang2026evoiragent,cui2026selfevolving,tang2026evolutionary,zhu2026epiagent,zhou2026hdragent} need to assemble restoration strategies on the fly for each input image, incurring repeated online planning overhead.
To address this problem, as shown in Fig.~\ref{fig:motiva}(b), implicit memory-based IR agents~\cite{chen2024restoreagent,lin2025jarvisir,li2026hybrid,jia2026tiragent,lu2026restorer1,yu2026derainagent,zhu2026opera,choi2026ditto,xiong2026beyond,zhang2026look,liu2025realworld} encode restoration strategy knowledge in model parameters through offline training, allowing restoration decisions to be directly obtained from model inference.
When adaptation to new IR tasks relies on MLLM fine-tuning, acquiring new restoration strategies changes parameters that also encode existing knowledge.
Such continual parameter adaptation is susceptible to catastrophic forgetting~\cite{zweiger2025selfadapting}, making knowledge retention a central concern when expanding an IR agent's capabilities.

To this end, we propose SIR-Agent, a skill-augmented IR agent that stores restoration strategy knowledge in an expandable IR skill library and keeps MLLM parameters frozen during adaptation.
Each task skill specifies applicability conditions, an ordered sequence of restoration operations, and tool configurations.
As shown in Fig.~\ref{fig:motiva}(c), given a degraded image, the agent progressively loads relevant skill information, selects a task skill, and executes its operations.
Since the final restored image alone cannot identify which operation should change, we introduce Chain Feedback-driven Skill Revision (CFSR) to revise task skills using feedback along their execution chains.
CFSR examines intermediate restoration states and chain feedback to locate underperforming operations, then retrieves and compares candidate tools to form a patched skill.
Because a revision that improves one input may degrade earlier cases, CFSR validates the patched skill on past degraded images and compares its outputs with those of the original skill.
The comparison informs the patched skill's applicability conditions and assesses compatibility through average quality gain and the rate of quality regression.
A compatible patch updates the original skill; otherwise, the original is retained and the patched skill is added as a new entry with its inferred applicability conditions.
To improve CFSR's revision efficiency, we further introduce a skill meta-learning mechanism driven by Recursive Self-Improvement (RSI)~\cite{wang2026metaskill}, which analyzes CFSR's revision trajectories to identify inefficient revision operations and autonomously refines its revision rules.
The updated rules guide subsequent skill revisions, whose trajectories in turn provide feedback for further rule refinement.

Our main contributions are as follows:
\begin{itemize}
    \item We propose SIR-Agent, which stores restoration strategy knowledge in an expandable skill library while keeping MLLM parameters frozen.
    Its CFSR mechanism uses chain feedback and replay-based compatibility assessment to revise existing skills or create new ones, enabling adaptation to new IR tasks while retaining restoration capabilities on known tasks.
    \item We present an RSI-driven skill meta-learning mechanism.
    It identifies inefficient operations in CFSR's revision trajectories and autonomously refines CFSR's revision rules, thereby improving the efficiency of subsequent skill revisions.
    \item Extensive experiments on 14 seen and 7 unseen IR benchmarks demonstrate that SIR-Agent achieves superior restoration performance against competing methods and strong generalization to new IR tasks.
\end{itemize}

%% file: sec/2_related.tex
\section{Related Work}
\label{sec:related}

\begin{figure*}[!t]
  \centering
  \includegraphics[width=\linewidth]{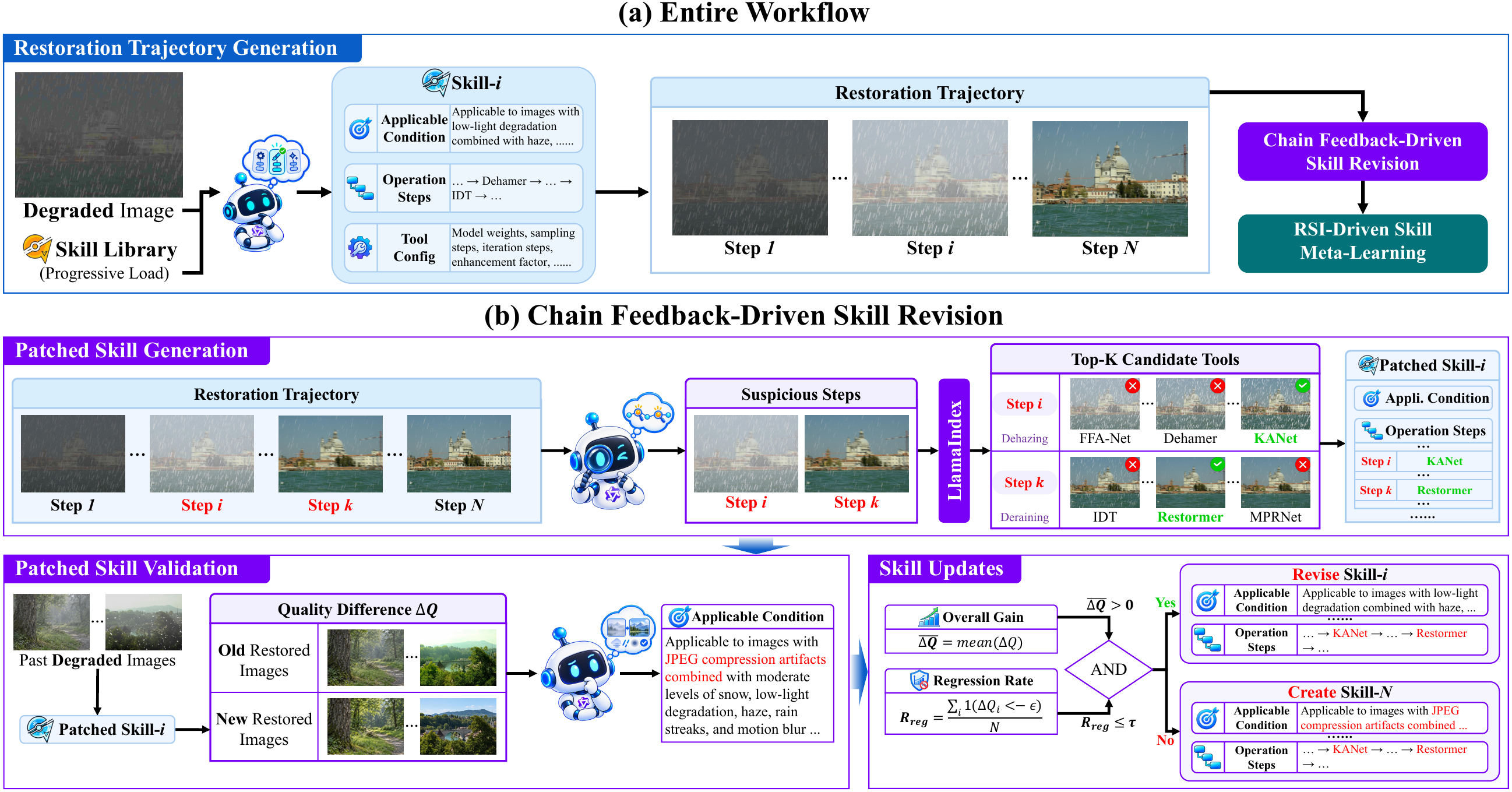}
  \caption{Overview of the proposed SIR-Agent architecture.
  (a) illustrates the overall workflow, where the frozen Qwen3.8-Max~\cite{qwen38} progressively loads relevant skills and executes the selected skill.
  The resulting restoration trajectories guide the CFSR to revise skills, while revision records guide the RSI-driven skill meta-learning to improve revision rules.
  (b) depicts the CFSR, which locates suspicious operation steps using chain feedback and uses LlamaIndex~\cite{liu2022llamaindex} to retrieve suitable candidate tools for skill revision. Replay on prior cases then assesses compatibility to determine whether to replace the original skill or add a new entry.}
  \label{fig:short-a}
\end{figure*}

\subsection{Image Restoration Agents}
\label{subsec:related_ir}

\textbf{Explicit Memory.} Explicit memory-based IR agents~\cite{zhu2025intelligent,jiang2026multiagent,zuo2025fourkagent,wang2026paagent,gao2026causalagentir,zhuang2026evoiragent,cui2026selfevolving,tang2026evolutionary,zhu2026epiagent,zhou2026hdragent} use different forms of externally stored restoration strategy knowledge to guide restoration decisions.
For task planning, external restoration strategy knowledge takes the form of experience documents and degradation-processing priors that guide task selection and execution order~\cite{zhu2025intelligent,jiang2026multiagent,zuo2025fourkagent}.
To capture context-dependent tool behavior, other methods organize external knowledge as tool portraits, causal relations, or hierarchical experience pools, supporting retrieval and feedback-driven updates~\cite{wang2026paagent,gao2026causalagentir,zhuang2026evoiragent}.
Beyond planning guidance, successful trajectories and coordination records preserve execution experience for reuse on subsequent inputs~\cite{cui2026selfevolving,tang2026evolutionary}.
Execution experience also informs specialized restoration tasks, where historical feedback is consolidated into statistical priors or retrievable cases to guide tool selection~\cite{zhu2026epiagent,zhou2026hdragent}.

Nevertheless, when retrieved knowledge cannot be directly reused for the current input, a restoration strategy must still be assembled online, incurring repeated planning overhead~\cite{zhu2025intelligent,gao2026causalagentir}.

\noindent
\textbf{Implicit Memory.} To reduce repeated online planning, implicit memory-based IR agents~\cite{chen2024restoreagent,lin2025jarvisir,li2026hybrid,jia2026tiragent,lu2026restorer1,yu2026derainagent,zhu2026opera,choi2026ditto,xiong2026beyond,zhang2026look,liu2025realworld} encode learned restoration knowledge in model parameters.
Learned policies and tool-effect models can shift part of the planning effort to training, reducing repeated online exploration~\cite{zhu2026opera,choi2026ditto,zhang2026look}.
Some methods train visual-language controllers using labeled data, tool-use instructions, and task feedback.
The controllers learn to select restoration tasks and tools based on the input degradations~\cite{chen2024restoreagent,lin2025jarvisir,li2026hybrid,jia2026tiragent}.
Lightweight controllers likewise learn policies for selecting restoration actions, tool paths, or expert configurations, encoding restoration strategy knowledge in their parameters~\cite{lu2026restorer1,yu2026derainagent,xiong2026beyond}.
Restoration decisions can additionally draw on learned models of tool effects, which support policy training through simulated trajectories or guide tool selection by predicting action outcomes before execution~\cite{choi2026ditto,zhang2026look}.
Parameter learning further extends from decision support to restoration execution, through joint optimization of plans and tools or quality-driven adaptation of weather-specific restorers alongside adaptive tool scheduling~\cite{zhu2026opera,liu2025realworld}.

However, fine-tuning an MLLM for new IR tasks may cause catastrophic forgetting of previously learned IR knowledge~\cite{zweiger2025selfadapting}.
In contrast, SIR-Agent keeps the MLLM parameters frozen and revises external IR skills to adapt to new tasks.
Moreover, our CFSR leverages intermediate chain feedback to locate weak operations and revise skills. 
It evaluates compatibility to choose between updating an existing skill and keeping it alongside a newly created one, making prior-task compatibility part of skill revision.

\subsection{Recursive Self-Improvement}
\label{subsec:related_rsi}

Recursive self-improvement (RSI) has attracted growing interest as an approach to agent optimization~\cite{zelikman2024selftaught,zhang2026hyperagents,wang2026metaskill}.
RSI aims to improve both task performance and the procedure that produces further improvements.
Agent optimization can target the external harness while keeping model parameters fixed~\cite{lee2026metaharness}.
The harness controls how the agent manages context, stores knowledge, and invokes tools.

At the code level, STOP~\cite{zelikman2024selftaught}, SICA~\cite{robeyns2025selfimproving}, and DGM~\cite{zhang2026darwin} revise their own programs and reuse them in later improvement rounds.
Hyperagents~\cite{zhang2026hyperagents} makes the improvement procedure directly editable as a meta-agent program.
For skill-based agents, MetaSkill-Evolve~\cite{wang2026metaskill} and Skill$^{\infty}$~\cite{tao2026skill} express the improvement procedure as editable meta-skills.
The meta-skills are themselves refined and then used to guide further skill optimization.

Building on RSI, we propose an RSI-driven skill meta-learning mechanism for SIR-Agent.
CFSR revises restoration skills using chain feedback.
Skill meta-learning then uses the revision trajectories to refine CFSR's rules for subsequent revisions.
Both restoration skills and their revision procedure therefore evolve while the MLLM parameters remain frozen.

%% file: sec/3_method.tex
\section{Method}
\label{sec:method}

\begin{figure*}[!t]
  \centering
  \includegraphics[width=\linewidth]{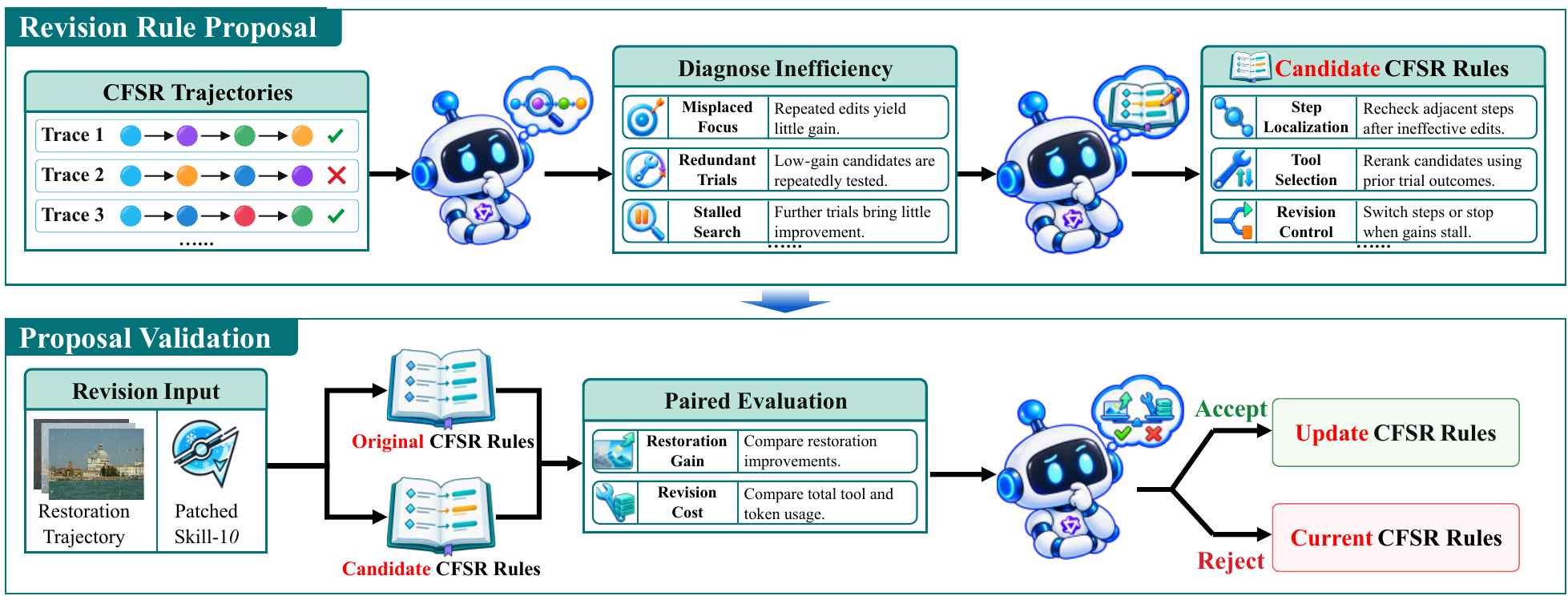}
  \caption{The illustration of the RSI-driven skill meta-learning.
  The MLLM analyzes revision trajectories to identify inefficient revision steps.
  The MLLM then proposes changes to the rules that guide CFSR's skill revision, such as those for step localization, tool selection, and revision control.
  The current and candidate rules are then evaluated on the same revision inputs.
  Paired comparisons of restoration gain and revision cost guide rule acceptance for subsequent skill revisions.}
  \label{fig:short-behaviours}
\end{figure*}

\subsection{Overview of SIR-Agent}
\label{subsec:skill_execution}

The overall architecture of SIR-Agent is shown in Fig.~\ref{fig:short-a}(a).
Given a degraded image, the SIR-Agent progressively loads relevant skills from the skill library.
The initial skill library is constructed from restoration trajectories, as detailed in Sec.~\ref{sec:experimental_setup}.
Each skill specifies applicability conditions, ordered operations, and tool configurations.
Once candidate skills are collected, the MLLM selects the skill whose applicability conditions match the observed degradations.
The SIR-Agent invokes the configured tools in the specified order to obtain the restored image.
The restoration trajectory provides CFSR with intermediate images, tool calls, configurations, and feedback to revise the selected skill.

In the skill revision stage of Fig.~\ref{fig:short-a}(a), CFSR generates a patched skill from restoration-chain feedback and validates it on prior cases before updating the skill library.
The revised skills support subsequent tasks, while revision trajectories record the update process.
RSI-driven skill meta-learning uses the revision trajectories to refine CFSR's rules for subsequent skill revisions.
The two modules are detailed in Secs.~\ref{subsec:cfsr} and~\ref{subsec:skill_meta_learning}.

\subsection{Chain Feedback-driven Skill Revision}
\label{subsec:cfsr}

Restoration skills enable strategy reuse but may become ineffective when degradation conditions change.
Relying only on the final restored image makes it difficult to identify which operation needs revision, because the image reflects the combined effects of all operations.
Therefore, we propose Chain Feedback-driven Skill Revision (CFSR), which uses restoration-chain feedback to guide targeted skill revision.
As shown in Fig.~\ref{fig:short-a}(b), CFSR comprises patched skill generation, patched skill validation, and skill updates.
Patched skill generation takes a skill \(s\) and its restoration trajectory as inputs.
The MLLM examines intermediate images and chain feedback to identify suspicious steps.
Localization considers residual degradations and the effects of each operation on subsequent steps.
Using the restoration trajectory and identified operations, CFSR retrieves the top-\(K\) candidate tools through LlamaIndex~\cite{liu2022llamaindex}.
For each identified operation, CFSR tests candidate tools on the same intermediate input image and compares the restoration quality of their outputs.
The best-performing tool is selected for each operation, and the selected replacements form a patched skill \(s'\).
The patched skill updates the associated tool configurations while retaining the operation order.

During patched skill validation, CFSR applies \(s'\) to historical degraded images \(\{u_n\}_{n=1}^{N}\) associated with \(s\).
The new restored images are compared with the corresponding old restored images obtained using \(s\).
Following the metric set used in \cite{wang2026paagent}, we evaluate the old and new restored images with CLIP-IQA~\cite{wang2023exploring}, HyperIQA~\cite{su2020hyperiqa}, NIQE~\cite{mittal2013making}, CPBD~\cite{narvekar2011noreference}, BRISQUE~\cite{mittal2012noreference}, MUSIQ~\cite{ke2021musiq}, and LIQE~\cite{zhang2023liqe}.
To combine the metric scores, we normalize them to a common scale, with NIQE~\cite{mittal2013making} and BRISQUE~\cite{mittal2012noreference} reversed so that higher values consistently indicate better quality.
The normalized scores are then aggregated into a single quality score \(Q\).
For each historical input \(u_n\), \(F_s(u_n)\) and \(F_{s'}(u_n)\) denote the old and new restored images, respectively.
The per-image quality change is:
\begingroup

\begin{equation}
    \Delta Q_n=Q(F_{s'}(u_n))-Q(F_s(u_n)).
    \label{eq:replay_difference}
\end{equation}
\endgroup
The MLLM uses the paired restoration results and quality differences to infer the patched skill's applicability conditions.

For skill updates, compatibility is assessed using the average quality gain \(\overline{\Delta Q}\) and regression rate \(R_{\mathrm{reg}}\):
\begingroup

\begin{equation}
    \overline{\Delta Q}=\frac{1}{N}\sum_{n=1}^{N}\Delta Q_n,
    \qquad
    R_{\mathrm{reg}}=\frac{1}{N}\sum_{n=1}^{N}
    \mathbf{1}[\Delta Q_n<-\epsilon],
    \label{eq:compatibility_metrics}
\end{equation}
\endgroup
where \(\epsilon\geq 0\) is a tolerance for small quality fluctuations and \(\mathbf{1}[\cdot]\) is the indicator function.
\(\overline{\Delta Q}\) measures the overall benefit, while \(R_{\mathrm{reg}}\) captures losses on individual prior cases.
However, a positive \(\overline{\Delta Q}\) can hide regressions on individual inputs.
Therefore, a patched skill is compatible when \(\overline{\Delta Q}>0\) and \(R_{\mathrm{reg}}\leq\tau\), where \(\tau\) is the permitted regression rate.
A compatible patch updates the original skill; otherwise, the original skill is retained and the patched skill is added as a new entry with its inferred applicability conditions.
CFSR records revision decisions, tool trials, replay outcomes, library updates, and resource usage in a revision trajectory, which RSI-driven skill meta-learning analyzes to improve CFSR's revision rules.

\subsection{RSI-Driven Skill Meta-Learning}
\label{subsec:skill_meta_learning}

Fixed revision rules cannot incorporate lessons from earlier revisions, so CFSR may repeat ineffective edits and redundant tool trials on similar cases.
Inspired by RSI~\cite{wang2026metaskill}, we design a skill meta-learning mechanism that improves CFSR's external revision instructions using recorded trajectories, as illustrated in Fig.~\ref{fig:short-behaviours}.
The MLLM examines the trajectories to connect inefficient revision steps with the instructions that guide them.
For example, repeated edits without quality gains motivate changes to step priorities.
Low-yield trials motivate adjustments to tool retrieval and ranking, while stalled search guides changes to candidate expansion, operation switching, or stopping.
The resulting proposals target external revision rules, including those for step localization, tool selection, and revision control.
To evaluate the proposals, CFSR is run with the current and candidate rules on the same validation cases and starting skills, using the same tool library and budget limits.
Restoration gain is the mean \(Q\) improvement over the starting skills; revision cost is measured by actual tool calls and MLLM token usage.
At round \(t\), the next CFSR rule set is selected from the current set \(\mathcal{R}_t^{\mathrm{cur}}\) and the candidate set \(\mathcal{R}_t^{\mathrm{cand}}\) proposed by the MLLM:
\begingroup
\begin{equation}
    \mathcal{R}_{t+1}^{\mathrm{cur}}=
    \begin{cases}
        \mathcal{R}_t^{\mathrm{cand}},
        & \mathcal{R}_t^{\mathrm{cand}}\in\mathcal{F}_t,\;
          \mathcal{R}_t^{\mathrm{cand}}\succ\mathcal{R}_t^{\mathrm{cur}},\\
        \mathcal{R}_t^{\mathrm{cur}}, & \text{otherwise},
    \end{cases}
    \label{eq:rule_update}
\end{equation}
\endgroup
where \(\mathcal{F}_t\) denotes the candidate rule sets that meet the predefined requirement for preserving prior-task performance after skill revision.
The relation \(\succ\) gives priority to higher restoration gain, with lower revision cost preferred when gains are comparable.
Equation~\ref{eq:rule_update} adopts the candidate rules only when both conditions hold; otherwise, the current rules are retained.
Section~\ref{subsec:cfsr} decides how to update individual skills, while Eq.~\ref{eq:rule_update} selects the rules that guide subsequent revisions.
Accepted rules guide subsequent CFSR runs, whose revision trajectories provide feedback for the next rule update.
The recursive feedback refines the revision procedure itself, aiming to reduce repeated ineffective edits and unnecessary tool trials.

%% file: sec/4_experiment.tex
\section{Experiments}
\label{sec:experiments}

\begin{table}[!t]
    \centering
    \caption{Quantitative comparison on deblurring (GoPro~\cite{nah2017deep} and HIDE~\cite{shen2019human}) and dehazing (SOTS~\cite{li2019benchmarking} and Dense-Haze~\cite{ancuti2019dense}). Each task reports mean PSNR and SSIM. Best results are in red.}
    \label{tab:main_comparison}
    \resizebox{\linewidth}{!}{%
        \begin{tabular}{c|c|cc|cc}
            \hline
            \rowcolor{irTableHeader} Method & Venue & \multicolumn{2}{c|}{\textbf{Deblurring}} & \multicolumn{2}{c}{\textbf{Dehazing}} \\
            \hline\hline
            PromptIR~\cite{potlapalli2023promptir} & NeurIPS'23 & 23.23 & 0.76 & 18.53 & 0.68 \\
            RAM~\cite{qin2024restore} & ECCV'24 & 26.89 & 0.85 & 15.67 & 0.62 \\
            CPLIR~\cite{wu2026beyond} & TPAMI'26 & 23.98 & 0.76 & 13.92 & 0.59 \\
            HOGformer~\cite{wu2026gradient} & AAAI'26 & 27.00 & 0.86 & 16.60 & 0.65 \\
            AgenticIR~\cite{zhu2025intelligent} & ICLR'25 & 23.77 & 0.84 & 15.10 & 0.65 \\
            4KAgent~\cite{zuo2025fourkagent} & NeurIPS'25 & 21.90 & 0.82 & 15.61 & 0.65 \\
            JarvisIR~\cite{lin2025jarvisir} & CVPR'25 & 21.49 & 0.79 & 14.22 & 0.60 \\
            OPERA~\cite{zhu2026opera} & NeurIPS'26 & 22.94 & 0.82 & 15.08 & 0.64 \\
            HarnessIR~\cite{kong2026harnessir} & arXiv'26 & 23.81 & 0.76 & 12.65 & 0.54 \\
            \hline
            SIR-Agent & -- & \textcolor{red}{31.83} & \textcolor{red}{0.95} & \textcolor{red}{26.12} & \textcolor{red}{0.81} \\
            \hline
        \end{tabular}%
    }
\end{table}

\newcommand{\ircomparisoncase}[1]{%
    \begingroup
    \setlength{\tabcolsep}{0.6pt}
    \renewcommand{\arraystretch}{0}
    \fontsize{9}{10.8}\selectfont
    \newcommand{\ircaseimage}[1]{\includegraphics[width=\dimexpr(1\textwidth-12\tabcolsep)/7\relax]{figs/Show/#1/##1}}
    \begin{tabular}{@{}*{7}{c}@{}}
        \ircaseimage{01_Raw} &
        \ircaseimage{06_AgenticIR.png} &
        \ircaseimage{07_4KAgent.png} &
        \ircaseimage{08_JarvisIR.png} &
        \ircaseimage{09_OPERA.png} &
        \ircaseimage{10_HarnessIR.png} &
        \ircaseimage{11_Our.png}
    \end{tabular}
    \endgroup
}

\newcommand{\ircomparisonlabels}{%
    \begingroup
    \setlength{\tabcolsep}{0.6pt}
    \fontsize{9}{10.8}\selectfont
    \newcommand{\irlabel}[1]{\makebox[\dimexpr(\textwidth-12\tabcolsep)/7\relax][c]{##1}}
    \begin{tabular}{@{}*{7}{c}@{}}
        \irlabel{Input} &
        \irlabel{AgenticIR~\cite{zhu2025intelligent}} &
        \irlabel{4KAgent~\cite{zuo2025fourkagent}} &
        \irlabel{JarvisIR~\cite{lin2025jarvisir}} &
        \irlabel{OPERA~\cite{zhu2026opera}} &
        \irlabel{HarnessIR~\cite{kong2026harnessir}} &
        \irlabel{Ours}
    \end{tabular}
    \endgroup
}

\begin{figure*}[!t]
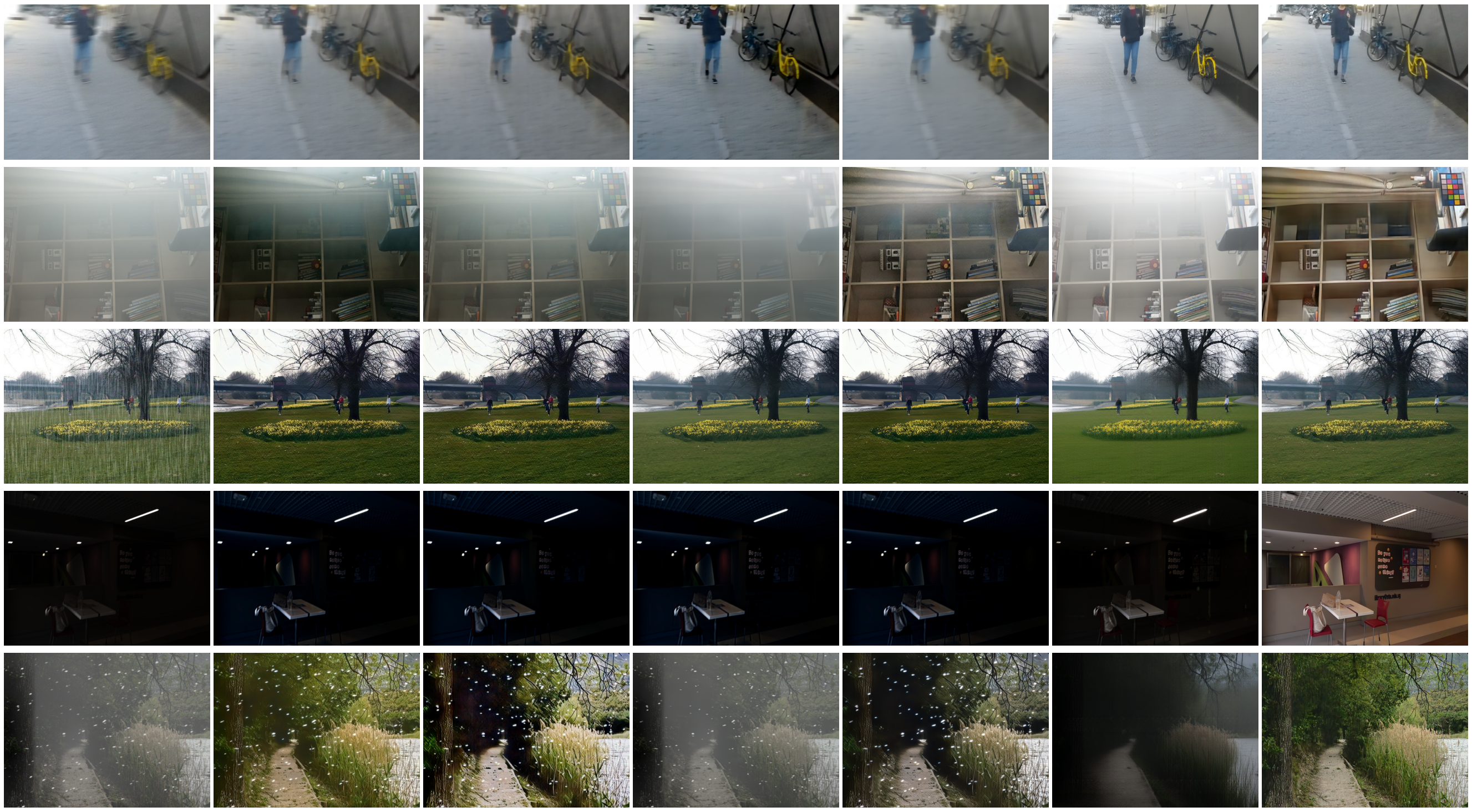

    \centering
    \ircomparisoncase{Image_Deblurring/HIDE/test_test-close-ups_4fromGOPR1087.MP4}\par\vspace{1.5pt}
    \ircomparisoncase{Image_Dehazing/Dense-Haze/hazy_37_hazy}\par\vspace{1.5pt}
    \ircomparisoncase{Image_Deraining/Rain13K/test_Test2800_Raw_802_5}\par\vspace{1.5pt}
    \ircomparisoncase{Compositional_Restoration/LOL-Blur/Raw_Blur_0232_0011}\par\vspace{1.5pt}
    \ircomparisoncase{Compositional_Restoration/CDD-11/low_haze_snow_00028}\par\vspace{0.8pt}
    \ircomparisonlabels
    \caption{Visual comparisons under various IR tasks. From top to bottom: deblurring, dehazing, deraining, low-light with blur, and CDR.}
    \label{fig:qualitative_comparison}
\end{figure*}

\setlength{\dbltextfloatsep}{8pt plus 2pt minus 2pt}

\begin{table*}[!t]
    \centering
    \begin{minipage}[t]{\columnwidth}
    \vspace{0pt}
\centering
    \caption{Quantitative comparison on denoising (BSD68~\cite{martin2001database} and SIDD~\cite{abdelhamed2018high}) and deraining (Rain13K~\cite{zamir2021multistage} and Rain1400~\cite{fu2017removing}). Each task reports mean PSNR and SSIM. Best results are in red.}
    \label{tab:main_comparison_noise_rain_snow}
    \resizebox{\linewidth}{!}{%
        \begin{tabular}{c|c|cc|cc}
            \hline
            \rowcolor{irTableHeader} Method & Venue & \multicolumn{2}{c|}{\textbf{Denoising}} & \multicolumn{2}{c}{\textbf{Deraining}} \\
            \hline\hline
            PromptIR~\cite{potlapalli2023promptir} & NeurIPS'23 & 28.24 & 0.82 & 24.49 & 0.79 \\
            RAM~\cite{qin2024restore} & ECCV'24 & 29.44 & 0.82 & 28.57 & 0.87 \\
            CPLIR~\cite{wu2026beyond} & TPAMI'26 & 31.64 & 0.85 & 26.55 & 0.85 \\
            HOGformer~\cite{wu2026gradient} & AAAI'26 & 22.56 & 0.75 & 24.00 & 0.76 \\
            AgenticIR~\cite{zhu2025intelligent} & ICLR'25 & 26.97 & 0.84 & 28.13 & 0.89 \\
            4KAgent~\cite{zuo2025fourkagent} & NeurIPS'25 & 25.52 & 0.81 & 26.79 & 0.87 \\
            JarvisIR~\cite{lin2025jarvisir} & CVPR'25 & 18.89 & 0.65 & 23.97 & 0.84 \\
            OPERA~\cite{zhu2026opera} & NeurIPS'26 & 22.44 & 0.80 & 27.43 & 0.88 \\
            HarnessIR~\cite{kong2026harnessir} & arXiv'26 & 23.81 & 0.72 & 19.79 & 0.58 \\
            \hline
            SIR-Agent & -- & \textcolor{red}{35.30} & \textcolor{red}{0.93} & \textcolor{red}{32.71} & \textcolor{red}{0.93} \\
            \hline
        \end{tabular}%
    }
    \end{minipage}\hfill
    \begin{minipage}[t]{\columnwidth}
    \vspace{0pt}
\centering
    \caption{Quantitative comparison on desnowing (CSD~\cite{chen2021allsnow} and RealSnow~\cite{zhu2023weather}), LLIE (LOL-V1~\cite{wei2018deepretinex} and LOL-V2~\cite{yang2021sparse}) and CDR (LOL-Blur~\cite{zhou2022lednet} and CDD-11~\cite{guo2025onerestore}). Each task reports mean PSNR and SSIM. Best results are in red.}
    \label{tab:main_comparison_lowlight_composite}
    \resizebox{\linewidth}{!}{%
        \begin{tabular}{c|cc|cc|cc}
            \hline
            \rowcolor{irTableHeader} Method & \multicolumn{2}{c|}{\textbf{Desnowing}} & \multicolumn{2}{c|}{\textbf{LLIE}} & \multicolumn{2}{c}{\textbf{CDR}} \\
            \hline\hline
            PromptIR~\cite{potlapalli2023promptir} & 22.40 & 0.80 & 8.86 & 0.23 & 12.82 & 0.46 \\
            RAM~\cite{qin2024restore} & 23.40 & 0.80 & 20.70 & 0.78 & 13.20 & 0.52 \\
            CPLIR~\cite{wu2026beyond} & 21.48 & 0.80 & 9.15 & 0.26 & 12.86 & 0.48 \\
            HOGformer~\cite{wu2026gradient} & 22.44 & 0.80 & 24.64 & 0.85 & 16.60 & 0.70 \\
            AgenticIR~\cite{zhu2025intelligent} & 21.86 & 0.75 & 11.80 & 0.44 & 13.98 & 0.58 \\
            4KAgent~\cite{zuo2025fourkagent} & 19.09 & 0.73 & 11.81 & 0.43 & 14.07 & 0.58 \\
            JarvisIR~\cite{lin2025jarvisir} & 18.85 & 0.75 & 11.88 & 0.44 & 14.26 & 0.58 \\
            OPERA~\cite{zhu2026opera} & 20.54 & 0.74 & 11.76 & 0.44 & 13.72 & 0.56 \\
            HarnessIR~\cite{kong2026harnessir} & 18.43 & 0.66 & 9.09 & 0.23 & 12.44 & 0.42 \\
            \hline
            SIR-Agent & \textcolor{red}{34.47} & \textcolor{red}{0.93} & \textcolor{red}{26.56} & \textcolor{red}{0.90} & \textcolor{red}{24.55} & \textcolor{red}{0.91} \\
            \hline
        \end{tabular}%
    }
    \end{minipage}
\end{table*}

\subsection{Experiment Settings}
\label{sec:experimental_setup}

\noindent\textbf{Implementation Details.}
To build the initial skill library, we collect restoration trajectories through exhaustive tool trials on the training set used by PaAgent~\cite{wang2026paagent}.
We retain complete trajectories whose feedback indicates successful restoration.
Following SkillIR~\cite{shao2026skillir}, we divide each retained trajectory into consecutive segments, each addressing one degradation.
The frozen Qwen3.8-Max~\cite{qwen38} uses a fixed prompt to summarize each segment's scene conditions and restoration effects from intermediate images and feedback.
We combine the segment summaries into applicability conditions for the complete restoration procedure.
Each trajectory then yields one skill containing the applicability conditions, ordered operations, and tool configurations.
Finally, we merge skills with compatible applicability conditions and execution procedures into the initial library.

\noindent\textbf{Compared Methods.}
We compare our SIR-Agent method with nine IR methods.
Four all-in-one IR methods are PromptIR~\cite{potlapalli2023promptir}, RAM~\cite{qin2024restore}, CPLIR~\cite{wu2026beyond}, and HOGformer~\cite{wu2026gradient}.
Five IR agents are AgenticIR~\cite{zhu2025intelligent}, 4KAgent~\cite{zuo2025fourkagent}, JarvisIR~\cite{lin2025jarvisir}, OPERA~\cite{zhu2026opera}, and HarnessIR~\cite{kong2026harnessir}.

\noindent\textbf{Benchmark Datasets.}
The main evaluation covers 14 IR benchmarks, grouped by IR task in Tables~\ref{tab:main_comparison}--\ref{tab:main_comparison_lowlight_composite}.
Deblurring uses GoPro~\cite{nah2017deep} and HIDE~\cite{shen2019human}, while dehazing uses SOTS~\cite{li2019benchmarking} and Dense-Haze~\cite{ancuti2019dense}.
Denoising is evaluated on BSD68~\cite{martin2001database} and SIDD~\cite{abdelhamed2018high}.
Deraining uses Rain13K~\cite{zamir2021multistage} and Rain1400~\cite{fu2017removing}, and desnowing uses CSD~\cite{chen2021allsnow} and RealSnow~\cite{zhu2023weather}.
Low-Light Image Enhancement (LLIE) is evaluated on LOL-V1~\cite{wei2018deepretinex} and LOL-V2~\cite{yang2021sparse}.
LOL-Blur~\cite{zhou2022lednet} and CDD-11~\cite{guo2025onerestore} cover Composite Degradation Restoration (CDR).
Generalization is evaluated on seven datasets unseen during initial skill construction.
These include NightIRS~\cite{fang2026depth} for dereflection, ISTD~\cite{wang2018stacked} for deshadow, and FHDMi~\cite{he2020fhde2net} for demoireing.
We also use LSMU~\cite{wang2026mcsfnet} for Underwater Image Enhancement (UIE), SID~\cite{huang2022sidnet} for dedusting, Flare7K++~\cite{dai2024flare7kpp} for Nighttime Flare Removal (NFR), and Smoke~\cite{jin2022structure} for desmoking.

\begin{table*}[!t]
    \centering
    \caption{Quantitative comparison on seven unseen tasks for dereflection (NightIRS~\cite{fang2026depth}), deshadow (ISTD~\cite{wang2018stacked}), demoireing (FHDMi~\cite{he2020fhde2net}), UIE (LSMU~\cite{wang2026mcsfnet}), dedusting (SID~\cite{huang2022sidnet}), NFR (Flare7K++~\cite{dai2024flare7kpp}), and desmoking (Smoke~\cite{jin2022structure}). Each task reports mean PSNR and SSIM. Best results are in red.}
    \label{tab:generalization}
    \resizebox{\linewidth}{!}{%
        \begin{tabular}{c|cc|cc|cc|cc|cc|cc|cc}
            \hline
            \rowcolor{irTableHeader} Method & \multicolumn{2}{c|}{\textbf{Dereflection}} & \multicolumn{2}{c|}{\textbf{Deshadow}} & \multicolumn{2}{c|}{\textbf{Demoireing}} & \multicolumn{2}{c|}{\textbf{UIE}} & \multicolumn{2}{c|}{\textbf{Dedusting}} & \multicolumn{2}{c|}{\textbf{NFR}} & \multicolumn{2}{c}{\textbf{Desmoking}} \\
            \hline\hline
            Input & 22.75 & 0.81 & 21.17 & 0.89 & 18.48 & 0.78 & 19.03 & 0.80 & 11.79 & 0.68 & 22.84 & 0.90 & 13.11 & 0.51 \\
            PromptIR~\cite{potlapalli2023promptir} & 22.34 & 0.78 & 19.13 & 0.87 & 17.68 & 0.73 & 18.82 & 0.81 & 12.24 & 0.72 & 22.59 & 0.87 & 13.28 & 0.52 \\
            RAM~\cite{qin2024restore} & 22.79 & 0.82 & 21.12 & 0.89 & 18.40 & 0.78 & 19.05 & 0.79 & 11.81 & 0.67 & 22.92 & 0.90 & 13.11 & 0.51 \\
            CPLIR~\cite{wu2026beyond} & 22.67 & 0.81 & 20.19 & 0.85 & 18.54 & 0.79 & 18.76 & 0.75 & 11.83 & 0.65 & 22.86 & 0.88 & 13.36 & 0.49 \\
            HOGformer~\cite{wu2026gradient} & 22.40 & 0.80 & 19.18 & 0.87 & 17.63 & 0.78 & 18.97 & 0.81 & 12.06 & 0.71 & 19.61 & 0.81 & 13.08 & 0.52 \\
            AgenticIR~\cite{zhu2025intelligent} & 17.98 & 0.74 & 21.17 & 0.89 & 18.11 & 0.76 & 17.20 & 0.77 & 13.09 & 0.79 & 18.91 & 0.79 & 13.04 & 0.52 \\
            4KAgent~\cite{zuo2025fourkagent} & 16.93 & 0.71 & 20.66 & 0.88 & 17.17 & 0.74 & 17.08 & 0.75 & 13.09 & 0.79 & 17.93 & 0.77 & 13.04 & 0.52 \\
            JarvisIR~\cite{lin2025jarvisir} & 17.64 & 0.69 & 16.07 & 0.75 & 17.13 & 0.70 & 17.11 & 0.76 & 13.10 & 0.70 & 17.66 & 0.73 & 12.64 & 0.50 \\
            OPERA~\cite{zhu2026opera} & 17.54 & 0.72 & 19.22 & 0.80 & 17.19 & 0.73 & 16.54 & 0.75 & 13.67 & 0.77 & 17.18 & 0.76 & 13.00 & 0.52 \\
            HarnessIR~\cite{kong2026harnessir} & 20.07 & 0.63 & 16.84 & 0.56 & 18.52 & 0.69 & 15.00 & 0.49 & 14.02 & 0.70 & 19.55 & 0.74 & 11.73 & 0.36 \\
            SIR-Agent & \textcolor{red}{24.18} & \textcolor{red}{0.83} & \textcolor{red}{30.55} & \textcolor{red}{0.92} & \textcolor{red}{21.82} & \textcolor{red}{0.86} & \textcolor{red}{23.41} & \textcolor{red}{0.86} & \textcolor{red}{24.63} & \textcolor{red}{0.88} & \textcolor{red}{29.06} & \textcolor{red}{0.94} & \textcolor{red}{16.00} & \textcolor{red}{0.57} \\
            \hline
        \end{tabular}%
    }
\end{table*}

\begin{figure*}[!t]
    \centering
    \begingroup
    \setlength{\tabcolsep}{0.6pt}
    \renewcommand{\arraystretch}{0.8}
    \ircomparisoncase{Image_Dereflecting/NightIRS/blended_0248}\par\vspace{1.5pt}
    \ircomparisoncase{Image_Deshadowing/ISTD/test_test_A_100-4}\par\vspace{1.5pt}
    \ircomparisoncase{UIE/LSMU/test_Raw_LSUI_1786_1_1}\par\vspace{1.5pt}
    \ircomparisoncase{Image_NFR/Flare7Kpp/test_data_real_input_input_000014}\par\vspace{1.5pt}
    \ircomparisoncase{Image_Desmoking/Smoke/test_hazy_33}\par\vspace{0.8pt}
    \ircomparisonlabels
    \endgroup
    \caption{Visual comparisons on unseen IR tasks. From top to bottom: dereflection, deshadow, UIE, NFR, and desmoking.}
    \label{fig:generalization}
\end{figure*}

\subsection{Comparison with SOTA Methods}
\label{sec:main_results}


\noindent\textbf{Quantitative Comparison.} Tables~\ref{tab:main_comparison}--\ref{tab:main_comparison_lowlight_composite} show that our SIR-Agent achieves the highest PSNR and SSIM~\cite{wang2004ssim} across all seven task groups.
On the six single-degradation tasks, our SIR-Agent improves PSNR by 1.92--11.07 dB over the strongest baseline for each task.
The advantage also holds for CDR: our SIR-Agent surpasses HOGformer~\cite{wu2026gradient} by 7.95 dB PSNR and 0.21 SSIM, demonstrating effective restoration when multiple degradations occur together.
The strongest PSNR baseline varies by task: HOGformer for deblurring, RAM for deraining, and CPLIR for denoising. 
SIR-Agent exceeds these baselines by 4.83, 4.14, and 3.66 dB, respectively.

\noindent\textbf{Qualitative Comparison.} In Fig.~\ref{fig:qualitative_comparison}, JarvisIR~\cite{lin2025jarvisir} leaves haze over the indoor scene, whereas our SIR-Agent reveals clearer shelves.
In the composite-degradation examples, AgenticIR~\cite{zhu2025intelligent}, 4KAgent~\cite{zuo2025fourkagent}, and OPERA~\cite{zhu2026opera} leave the furniture dark or retain snow along the path.
Our SIR-Agent reveals the furniture and suppresses haze and snow along the path.
More visual comparisons are provided in the supplementary material.

\newcommand{\ablationpanel}[2]{%
    \includegraphics[width=\dimexpr(\linewidth-6\tabcolsep)/4\relax]{figs/Show/#1/#2}%
}

\setcounter{topnumber}{1}
\setlength{\textfloatsep}{8pt plus 2pt minus 2pt}
\begin{table}[t]
\centering
    \caption{Effectiveness of CFSR on desmoking (new task) and CDR (prior task). Each task reports mean PSNR and SSIM. Tool calls and tokens are counted per skill revision.}
    \label{tab:ablation}
    \resizebox{\linewidth}{!}{%
        \begin{tabular}{c|cc|cc|c|c}
            \hline
            \rowcolor{irTableHeader} Variant & \multicolumn{2}{c|}{\textbf{New Task}} & \multicolumn{2}{c|}{\textbf{Prior Task}} & Tool calls & Tokens (k) \\
            \hline\hline
            \textit{w/ FS} & 13.45 & 0.52 & \textcolor{red}{24.55} & \textcolor{red}{0.91} & -- & -- \\
            \textit{w/o RR} & \textcolor{red}{15.62} & \textcolor{red}{0.56} & 23.88 & 0.89 & \textcolor{red}{19.60} & \textcolor{red}{24.80} \\
            \textit{w/ FR} & \textcolor{red}{15.62} & \textcolor{red}{0.56} & 24.51 & \textcolor{red}{0.91} & \textcolor{red}{19.60} & \textcolor{red}{24.80} \\
            \hline
        \end{tabular}%
    }
\par\vspace{12pt}
\centering
    \caption{Effectiveness of chain feedback on desmoking (new task) and CDR (prior task). Each task reports mean PSNR and SSIM. Tool calls and tokens are counted per skill revision. }
    \label{tab:ablation_feedback}
    \resizebox{\linewidth}{!}{%
        \begin{tabular}{c|cc|cc|c|c}
            \hline
            \rowcolor{irTableHeader} Variant & \multicolumn{2}{c|}{\textbf{New Task}} & \multicolumn{2}{c|}{\textbf{Prior Task}} & Tool calls & Tokens (k) \\
            \hline\hline
            \textit{w/ RF} & 14.78 & 0.54 & 24.49 & \textcolor{red}{0.91} & 26.80 & 29.60 \\
            \textit{w/o SSL} & 15.10 & 0.55 & 24.43 & 0.90 & 31.20 & 33.60 \\
            \textit{w/ FR} & \textcolor{red}{15.62} & \textcolor{red}{0.56} & \textcolor{red}{24.51} & \textcolor{red}{0.91} & \textcolor{red}{19.60} & \textcolor{red}{24.80} \\
            \hline
        \end{tabular}%
    }
\par\vspace{12pt}
\centering
    \caption{Effectiveness of the RSI strategy on desmoking (new task) and CDR (prior task). Each task reports mean PSNR and SSIM. Tool calls and tokens are counted per skill revision. }
    \label{tab:ablation_efficiency}
    \resizebox{\linewidth}{!}{%
        \begin{tabular}{c|cc|cc|c|c}
            \hline
            \rowcolor{irTableHeader} Variant & \multicolumn{2}{c|}{\textbf{New Task}} & \multicolumn{2}{c|}{\textbf{Prior Task}} & Tool calls & Tokens (k) \\
            \hline\hline
            \textit{w/ FR} & 15.62 & 0.56 & 24.51 & \textcolor{red}{0.91} & 19.60 & 24.80 \\
            \textit{w/o CP} & 15.96 & \textcolor{red}{0.57} & 24.53 & \textcolor{red}{0.91} & 18.40 & 23.50 \\
            \textit{full model} & \textcolor{red}{16.00} & \textcolor{red}{0.57} & \textcolor{red}{24.55} & \textcolor{red}{0.91} & \textcolor{red}{13.40} & \textcolor{red}{17.10} \\
            \hline
        \end{tabular}%
    }
\end{table}

\begin{figure}[t]
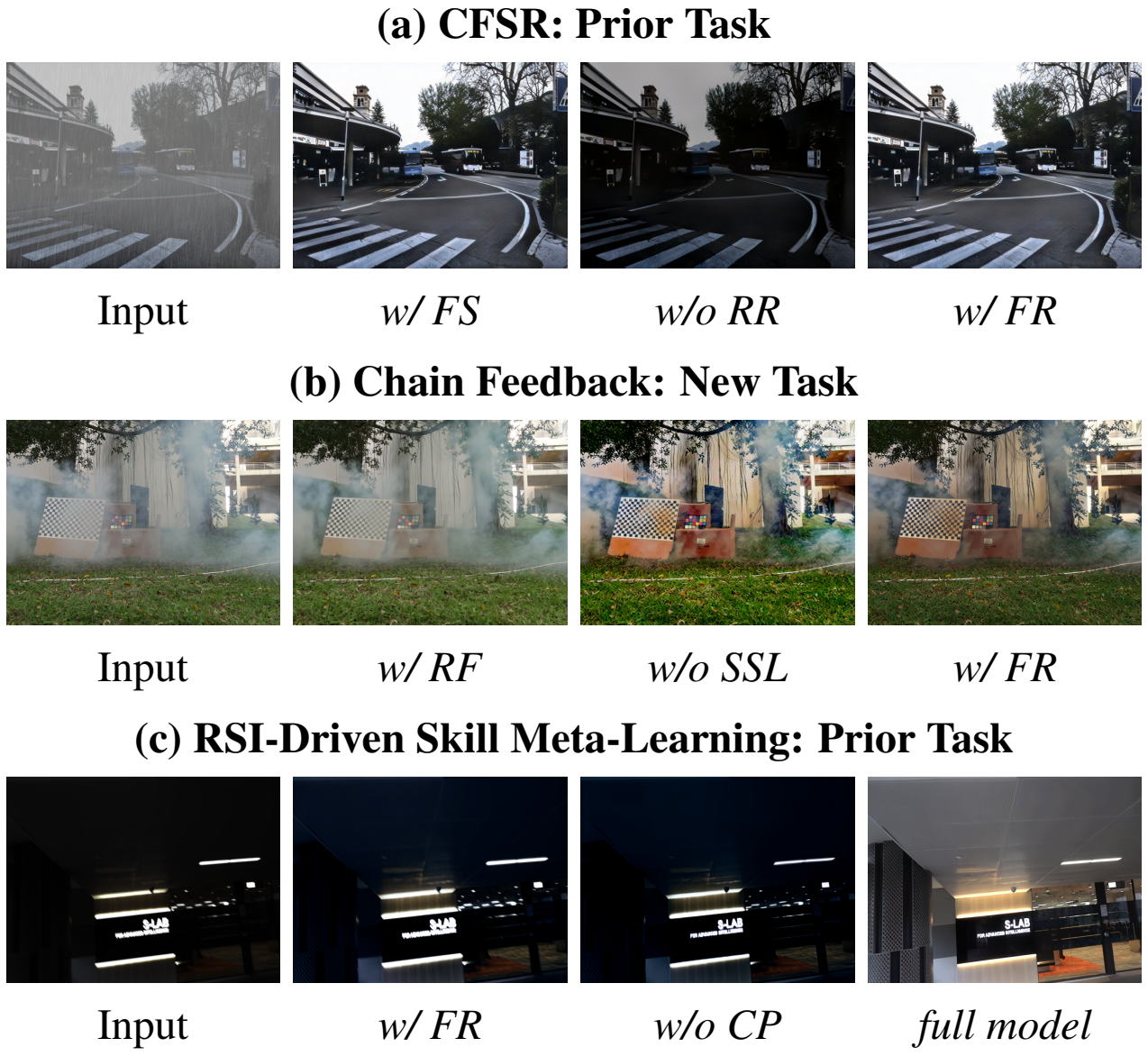

\centering
    \begingroup
    \setlength{\tabcolsep}{1.5pt}
    \fontsize{9}{10.8}\selectfont
    \begin{tabular}{@{}*{4}{c}@{}}
        \multicolumn{4}{c}{\textbf{(a) CFSR: Prior Task}} \\[2pt]
        \ablationpanel{Ablation/CFSR}{Input.png} &
        \ablationpanel{Ablation/CFSR}{w_FS.png} &
        \ablationpanel{Ablation/CFSR}{wo_RR.png} &
        \ablationpanel{Ablation/CFSR}{w_FR.png} \\[1pt]
        Input & \textit{w/ FS} & \textit{w/o RR} & \textit{w/ FR} \\[4pt]
        \multicolumn{4}{c}{\textbf{(b) Chain Feedback: New Task}} \\[2pt]
        \ablationpanel{Ablation/ChainFeedback}{Input.png} &
        \ablationpanel{Ablation/ChainFeedback}{w_RF.png} &
        \ablationpanel{Ablation/ChainFeedback}{wo_SSL.png} &
        \ablationpanel{Ablation/ChainFeedback}{w_FR.png} \\[1pt]
        Input & \textit{w/ RF} & \textit{w/o SSL} & \textit{w/ FR} \\[4pt]
        \multicolumn{4}{c}{\textbf{(c) RSI-Driven Skill Meta-Learning: Prior Task}} \\[2pt]
        \ablationpanel{Ablation/RSI}{Input.png} &
        \ablationpanel{Ablation/RSI}{w_FR.png} &
        \ablationpanel{Ablation/RSI}{wo_CP.png} &
        \ablationpanel{Ablation/RSI}{full_model.png} \\[1pt]
        Input & \textit{w/ FR} & \textit{w/o CP} & \textit{full model}
    \end{tabular}
    \endgroup
    \caption{Visual comparison of ablation configurations for (a) CFSR on the prior task, (b) chain feedback on the new task, and (c) the RSI-driven skill meta-learning on the prior task.}
    \label{fig:ablation}
\end{figure}

\subsection{Generalization to Unseen IR Tasks}
\label{sec:generalization}


\noindent\textbf{Quantitative Comparison.}
Our SIR-Agent obtains the highest PSNR and SSIM on all seven unseen tasks in Table~\ref{tab:generalization}.
Compared with the strongest baselines, our SIR-Agent gains 9.38 dB PSNR over AgenticIR~\cite{zhu2025intelligent} on deshadow and 2.64 dB over CPLIR~\cite{wu2026beyond} on desmoking.
Relative to the degraded inputs, SIR-Agent improves both metrics across all unseen tasks, with PSNR gains ranging from 1.43 dB for dereflection to 12.84 dB for dedusting.
The results show that our SIR-Agent adapts to new degradations through skill revision without MLLM fine-tuning.

\noindent\textbf{Qualitative Comparison.} 
In Fig.~\ref{fig:generalization}, our SIR-Agent removes reflections and shadows more effectively than AgenticIR~\cite{zhu2025intelligent} and 4KAgent~\cite{zuo2025fourkagent}, while preserving pavement patterns.
For UIE, our SIR-Agent produces outputs with a weaker green cast than those of JarvisIR~\cite{lin2025jarvisir} and HarnessIR~\cite{kong2026harnessir}.
For NFR, our SIR-Agent produces a smaller halo and clearer windows than OPERA~\cite{zhu2026opera}.
For desmoking, JarvisIR and HarnessIR retain smoke over the path and vegetation, whereas our SIR-Agent removes more smoke and reveals clearer foliage.
More visual comparisons are provided in the supplementary material.

\subsection{Ablation Study}
\label{sec:ablation}

We conduct an ablation study on CFSR, chain feedback, and the RSI-driven skill meta-learning to evaluate the contribution of each component.
The following configurations are compared:
\begin{itemize}
    \interlinepenalty=10000
    \item \textit{w/ FS}: Use a fixed skill library.
    \item \textit{w/ FR}: Update skills using fixed rules.
    \item \textit{w/o RR}: Remove the regression rate from CFSR.
    \item \textit{w/ RF}: Use result-level feedback for suspicious-step localization.
    \item \textit{w/o SSL}: Disable suspicious-step localization and try revising steps one by one in their original order.
    \item \textit{w/o CP}: Remove cost preference from the RSI-driven skill meta-learning.
    \item \textit{full model}: Using our full SIR-Agent.
\end{itemize}

\noindent\textbf{Effectiveness of CFSR.} 
The comparison between \textit{w/ FS} and \textit{w/ FR} in Table~\ref{tab:ablation} evaluates skill revision on both new and prior tasks.
Skill revision raises new-task PSNR by 2.17 dB over the fixed library, while prior-task PSNR changes from 24.55 to 24.51 dB.
The additional comparison with \textit{w/o RR} examines the regression-rate constraint.
Removing this constraint leaves both new-task metrics unchanged but lowers prior-task PSNR by 0.63 dB.
Figure~\ref{fig:ablation}(a) compares the corresponding prior-task restorations. Skill revision thus improves new-task restoration, while the regression-rate constraint helps retain prior-task performance.

\noindent\textbf{Effectiveness of Chain Feedback.} 
In Table~\ref{tab:ablation_feedback}, we compare \textit{w/ RF} with \textit{w/ FR} to assess chain feedback, and \textit{w/o SSL} with \textit{w/ FR} to assess step localization.
Compared with result-level feedback, chain feedback gains 0.84 dB on the new task while using 7.20 fewer tool calls per revision. Token usage also falls from 29.60k to 24.80k for this comparison.
Disabling step localization lowers new-task PSNR by 0.52 dB and increases tool calls by 11.60.
Figure~\ref{fig:ablation}(b) shows the corresponding outputs on the new task. Both comparisons show that chain feedback and step localization improve restoration quality with fewer tool calls.

\noindent\textbf{Effectiveness of the RSI-driven Skill Meta-learning.} 
Table~\ref{tab:ablation_efficiency} compares \textit{w/ FR} with the \textit{full model} to evaluate learned revision rules, and includes \textit{w/o CP} to examine cost preference.
The full model reduces tool calls and token usage by 31.6\% and 31.0\% relative to fixed rules, while improving new-task PSNR by 0.38 dB.
Removing cost preference adds 5.00 tool calls per revision and lowers new-task PSNR by 0.04 dB.
Figure~\ref{fig:ablation}(c) compares the prior-task outputs. The learned rules improve restoration and efficiency, with cost preference further reducing tool usage.

%% file: sec/5_conclusion.tex
\section{Conclusion}
\label{sec:conclusion}

In this paper, we present SIR-Agent, a skill-augmented IR agent designed to adapt to new IR tasks without MLLM fine-tuning.
The proposed CFSR uses restoration-chain feedback to locate weak operations and compatibility assessment to guide skill updates for new tasks while preserving prior restoration knowledge.
Meanwhile, we design an RSI-driven skill meta-learning mechanism that analyzes revision trajectories to identify ineffective edits and refine CFSR's revision rules.
The updated rules guide subsequent revisions, allowing skills and their revision strategy to evolve together.
Experiments on 14 benchmarks and 7 unseen datasets demonstrate SIR-Agent's superior restoration performance and generalization to new IR tasks.

\section*{LLM Usage Statement}
OpenAI Codex assisted with manuscript drafting and editing, literature checks, LaTeX preparation, and experiment planning.